\documentclass[]{article}
\usepackage{proceed2e}

\usepackage{times}
\usepackage{graphicx} 
\usepackage{subfigure} 
\usepackage{amsmath,amsthm}
\usepackage{slashbox}
\usepackage{multirow}
\usepackage{amssymb}

\usepackage{hyperref}

\title{Generalized K-fan Multimodal Deep Model with Shared Representations}

\author{} 

\author{ {\bf Gang Chen} and {\bf Sargur N. Srihari}  \thanks{Footnote for author to give an
alternate address.} \\
Computer Science Dept. \\
University at Buffalo, SUNY\\
Buffalo, NY 14260 \\
}

\begin{document}

\maketitle

\begin{abstract}
Multimodal learning with deep Boltzmann machines (DBMs) is an generative approach to fuse multimodal inputs, and can learn the shared representation via Contrastive Divergence (CD) for classification and information retrieval tasks. However, it is a 2-fan DBM model, and cannot effectively handle multiple prediction tasks. Moreover, this model cannot recover the hidden representations well by sampling from the conditional distribution when more than one modalities are missing. In this paper, we propose a K-fan deep structure model, which can handle the multi-input and muti-output learning problems effectively. In particular, the deep structure has K-branch for different inputs where each branch can be composed of a multi-layer deep model, and a shared representation is learned in an discriminative manner to tackle multimodal tasks. Given the deep structure, we propose two objective functions to handle two multi-input and multi-output tasks: joint visual restoration and labeling, and the multi-view multi-calss object recognition tasks. To estimate the model parameters, we initialize the deep model parameters with CD to maximize the joint distribution, and then we use backpropagation to update the model according to specific objective function. 
The experimental results demonstrate that the model can effectively leverages multi-source information and predict multiple tasks well over competitive baselines.
\end{abstract}

\section{Introduction}
We are exploring the multimodal learning in a joint framework when we have multiple forms of data available in the information age, such as images, labels, texts and videos. Each modality is characterized by very distinct statistical properties, but it also reflects one or two facets of the data even though they come from different input channels. Thus, it is possible to leverage different inputs to learn a shared representation in the prediction tasks, such as data restoration and classification. Recent advances in deep learning \cite{Hinton06b} and multi-modality learning \cite{Ngiam10} shed lights on joint representation learning which captures the real-world concept that the data corresponds to. 
The deep learning methods \cite{Hinton06a,Bengio12}, such as deep belief networks (DBNs) \cite{Hinton06a}, CNN \cite{Fukushima80,LeCun89} and LSTM \cite{Hochreiter97}, can learn an abstract and expressive representations, which can capture a huge number of possible input configurations. Hence, the representation learned is useful for classification and information retrieval. The multimodal learning model \cite{Srivastava14b} in a sense extends the deep learning framework, such as deepautoencoder or deep Boltzmann machines (DBMs), to handle different modalities. Thus it can learn a joint representation such that similarity in the code space implies similarity of the corresponding concepts. However, these previous multi-modality models \cite{Ngiam10,Srivastava14b} are kind of 2-fan deep model, and can only handle or predict one task. Often, the joint representation learned is not robust enough when the data is typically very noisy and there may be missing. Furthermore, how to leverage multi-source information from multimodalities is also an interesting topic for classification and information retrieval.  

In this paper, we propose a K-fan deep structure model, where we generalize the previous 2-fan multimodal learning \cite{Ngiam10,Srivastava14b} to handle multiple inputs and outputs. Our model is composed of K-branch deep models, with a shared hidden layer. In particular, the deep structure model has K pathway for different inputs respectively, and learn a shared representation to tackle multimodal tasks in an discriminative manner. Our model is powerful because each branch can be a multi-layer deep model, for example, we can use DBN, CNN or LSTM in each branch to handle different modalities, such as images, texts and videos. Refer to the right panel in Fig. \ref{fig:dbnkfan} with DBNs used in each branch for a 3-fan deep structure case. Most similar to our work are the bi-modal deep models \cite{Ngiam10,Srivastava14b} to handle image-text data or speech-vision data. The multimodal DBM proposed by Ngiam et al. {\cite{Ngiam10} used a deep autoencoder for speech and vision fusion, while the other method \cite{Srivastava14b} leveraged DBMs to learn hidden representations for bi-modal image-text data. There are, however, several crucial differences between our model and other methods. First, in this work we can handle the K-fan different modalities, instead of bi-modal inputs. Moreover, each branch can be a deep model, such as DBN, CNN and LSTM to handle different multimodalities. Thus, our model is more powerful than 2-fan modality models. Secondly, our deep structure can jointly learn from multiple inputs to predict multi-output with shared representations. Lastly, given the deep K-fan structure, it is very flexible to design an objective function and learn the model parameters in an discriminative manner by fusing multiple inputs. 


In this paper, we use a composition of multiple DBNs as a special case in our K-fan model. Note that other deep model such as CNN and LSTM can be used in each branch to handle more complex data. In the learning stage, all inputs are thought as observed data. Thus, we pretrain the model by leveraging all channels to learn the joint representations via Contrastive Divergence (CD). Then, we fine-tune the model parameters via backpropagation according to different tasks. In the inference stage, our model uses the feed-forward to handle multiple inputs and output predictions. 

\begin{figure}[t!]
\centering
\includegraphics[trim = 25mm 55mm 15mm 55mm, clip, width=8.6cm]{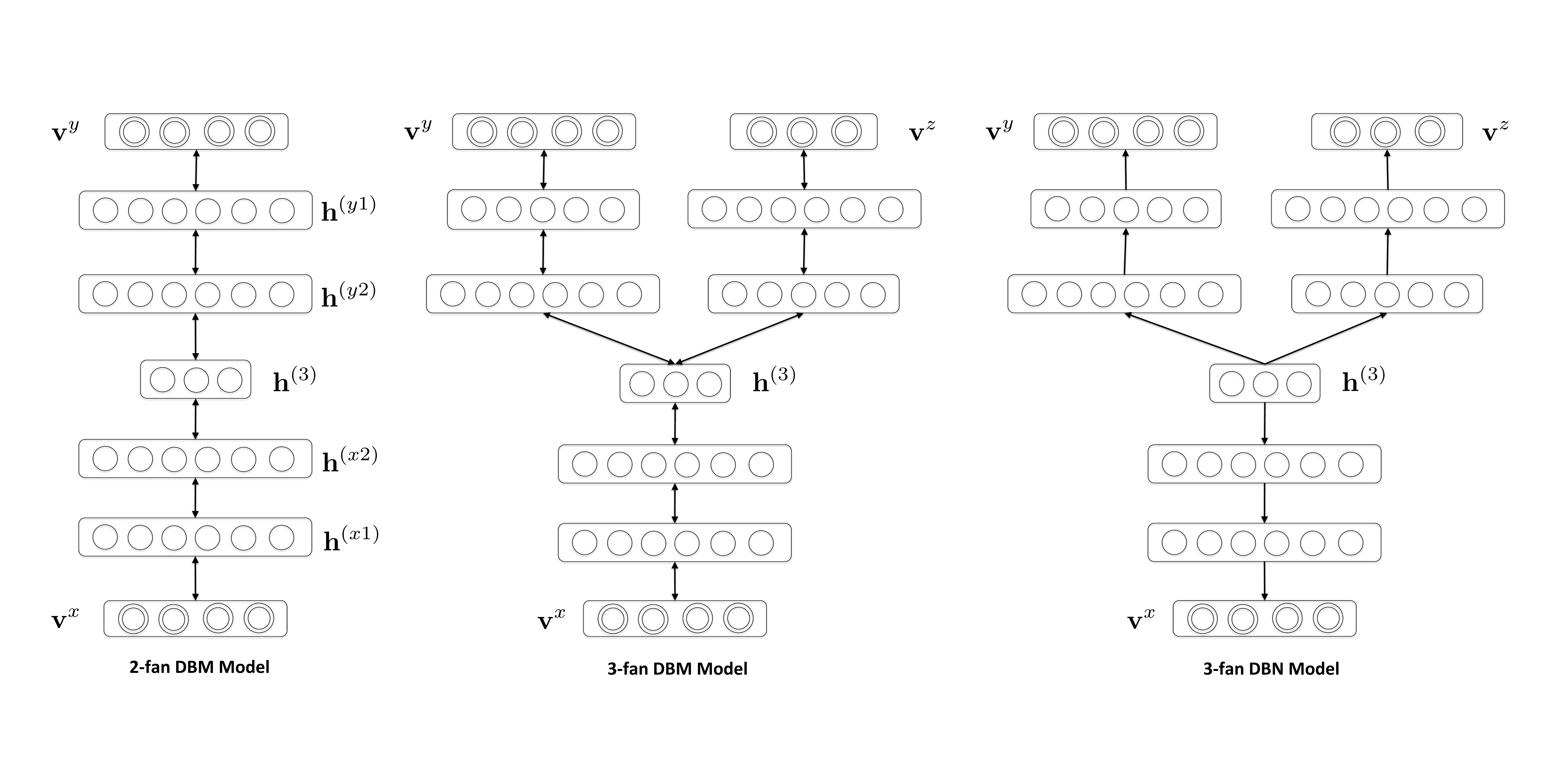}
\caption{The left is an example of 2-fan multimodal DBM, with shared representations via the 3-layer deep architecture. The middle is a 3-fan DBM, with a joint representation shared by ${\bf v}^x$, ${\bf v}^y$ and ${\bf v}^z$. The right is a 3-fan DBN, which is different from DBM and can be easily generalized into K-fan deep structure by extending more branches from the shared representations. Note that except the shared hidden nodes, each branch can be different deep models, such as DBN, CNN and LSTM, with different number of layers and also different number of nodes in each layer. }
\label{fig:dbnkfan}
\end{figure}

We test our model on two different tasks: joint visual restoration and labeling, as well as multiclass object recognition task. As for the first experiment, it is a multi-task learning problem, which need to jointly restore the data as well as label it. While for the second experiment, we leverage multiple inputs or resources to improve the prediction accuracy. The experimental results demonstrate the advantages of our model over other competitive baselines, such as multimodal DBM and support vector machines (SVM). 
\section{Related work}
Over the past few years, there have been several approaches proposed to learning from multimodal data. For example, a joint model of images and text using dual-wing harmoniums is builded by Xing et al. \cite{Xing05}, which is a generative model and can be viewed as a linear RBM model with Gaussian and Poisson visible units. 
Huiskes et al. \cite{Huiskes10} used standard low-level image features with additional captions, or tags, to improve classification accuracy significantly over SVM. A similar approach \cite{Guillaumin10}, based on multiple kernel learning framework, was also proposed and demonstrated that an additional text modality can improve the accuracy of SVMs on various object recognition tasks.

Recently, the multimodal learning with deep learning has attracted great attention in machine learning community.  The approach of Ngiam et al. \cite{Ngiam10}  used a deep autoencoder for speech and vision fusion. And a multimodal DBM \cite{Srivastava14b} is also proposed, which can be viewed as a composition of unimodal undirected pathways. Each pathway can be pretrained separately in a completely unsupervised fashion, with a large supply of unlabeled data. Any number of pathways each with any number of layers could potentially be used. 
One advantage of the multimodal DBM is that it is a generative model, which allows the model to naturally handle missing data in one channel by sampling. However, it cannot effectively handle data missing in multiple channels. More specifically, it cannot be used to predict multiple tasks. 
 
In this paper, we propose a unified K-fan deep neural network, which can learn a shared representation to handle different tasks. 
Moreover, each branch can be composed of a multi-layer deep model, such as CNN, DBN and LSTM to handle different inputs. Thus, our model is more powerful by generalizing previous multimodal models \cite{Ngiam10,Srivastava14b} for more complex tasks. We initialize the model parameters in a generative manner with CD algorithm. While in fine-tuning stage, we can update the model parameters by optimizing the given objective functions. Thus, our model can leverage multiple resources and also handle or predict multiple outputs. 
We test our model on two different tasks: joint visual restoration and recognition, and multi-view multi-class object recognition. 

As for the former task, usually, the visual restoration and recognition were addressed in separated pipeline, for example image denoising followed with recognition.  One related work is denoising autoencoder \cite{Vincent10}, which extends the work \cite{Holmstrom90,Hinton06b} and minimizes the reconstruction loss between the input (the corrupted data) and the output (the clean version of it), to learn feature representation. 
Recently, a robust Boltzmann machine (RoBM) \cite{Tang12} was introduced for recognition and denoising. This model added another shape RBM to the Gaussian RBM prior to model the noise variables which indicate where to ignore the occluder in the image. Thus, the RoBM contains a multiplicative gating mechanism to handle unexpected corruptions of the observed variables. However, the experiments only show its effectiveness for regular structural noise. 

As for the multi-view multi-class object recognition, Torralba et al. \cite{Torralba07} proposed the joint boosting method to learn shared representations for multi-view multi-class object detection. Huiskes et al. \cite{Huiskes10} used SVMs to leverage the additional view information to boost the classification performance. 

We compared our method to other competitive baselines such as multimodal DBM and SVMs, and the experimental results show the advantages of our model over a variety of tasks. 

\section{K-fan multimodal deep model}
Our K-fan deep structure is composed of K deep models in each branch, with a shared representation to handle multi-modalities. For clarity, we use the DBNs as the deep model in each branch in our K-fan deep structure to explain our model. More specifically, we use restricted Boltzmann machines (RBMs) as the building blocks in each branch. 
In the pretraining stage, we initialize the parameters with CD algorithm as the deep Boltzmann machines. In the fine-tuning stage, our model is more like the generalized deep autoendcoder with multiple pathways. In the following parts, we will review RBMs first, and then we will introduce our model.
\subsection{Background}
An RBM with $n$ hidden units is a parametric model of the joint distribution between a layer of hidden variables ${\bf h} \in \{0, 1\}^n$ and the observation ${\bf v} \in \{{0,1}\}^D$. The RBM joint likelihood takes the form:
\begin{equation}\label{eq:eq1}
p({\bf v},{\bf h}) \propto e^{-E({\bf v}, {\bf h})}
\end{equation}
where the energy function is 
\begin{equation}\label{eq:eq2}
E({\bf v},{\bf h}) = -{\bf h}^T{\bf W}{\bf v} - {\bf b}^T {\bf v} - {\bf c}^T{\bf h} 
\end{equation}
And we can compute the following conditional likelihood:
\begin{subequations}\label{eq:eq3}
\begin{align}
        & p({\bf v} | {\bf h}) = \prod_{i} p( v_{i} | {\bf h})\\
        & p(v_i=1 | {\bf h}) = \mathrm{logistic}(b_{i} + \sum_{j} W(i,j) h_j)\\
        & p(h_i =1 | {\bf v}) = \mathrm{logistic}(c_{i} + \sum_{j} W(j, i) v_j)
\end{align}
\end{subequations}
where $\mathrm{logistic}(x)  = 1/(1+e^{-x})$. To learn RBM parameters, we need to minimize the negative log likelihood $-\mathrm{log}p({\bf v})$ on training data, the parameters updating can be calculated with an efficient stochastic descent method, namely contrastive divergence (CD) \cite{Hinton06a}. Thus, we get the following stochastic gradient for ${\bf W}$ from CD,
\begin{align}
\frac{\partial{\mathrm{log}p({\bf v})}}{\partial{W_{ij}}} = \langle v_{i}h_{j}\rangle_{data} -  \langle v_{i}h_{j} \rangle_{model}  \label{eq:grad2} 
\end{align}
where we ignore the biases of both hidden and observation layers. 
And update $\theta$ until convergence with gradient descent
\begin{equation}
\theta = \theta + \eta \frac{\partial \mathrm{log} p({\bf v})}{\partial \theta}
\label{eq:grbm}
\end{equation}
where $\theta$ is the weight and biases, and $\eta$ is the learning rate. 

A deep (restricted) Boltzmann machine (DBM) is a stack of RBMs, in which each layer can capture complicated and higher-order correlations between the activities of hidden features in the layer below \cite{Salakhutdinov12}. For a two layer DBM, it contains one layer visible units ${\bf v} \in \{{0,1}\}^D$ and two layer hiddens variables ${\bf h}^{(1)} \in \{0, 1\}^{n_1}$ and ${\bf h}^{(2)} \in \{0, 1\}^{n_2}$. The energy of the joint configuration $\{ {\bf v}, {\bf h}^{(1)},  {\bf h}^{(2)} \}$ is defined as (ignoring bias terms):
\begin{equation}\label{eq:dbme}
E ({\bf v},{\bf h}; \theta) = -{\bf v}^T {\bf W}^{(1)} {\bf h}^{(1)} - {\bf h}^{{(1)}^{T}} {\bf W}^{(2)} {\bf h}^{(2)}
\end{equation}
where ${\bf h} = \{  {\bf h}^{(1)} , {\bf h}^{(2)}   \}$ represent the set of hidden units, and $\theta = \{{\bf W}^{(1)}, {\bf W}^{(2)}  \}$ are model parameters,  representing visible-to-hidden and hidden-to-hidden symmetric interaction terms. Similar to RBMs, this binary-binary DBM can be easily extended to modeling dense real-valued or sparse count data, which has been extensively discussed in \cite{Srivastava14b}. Then, we can get the following likelihood by marginalizing out ${\bf h}$
\begin{align}\label{eq:dbm}
&P({\bf v}; \theta) = \sum_{{\bf h}^{(1)} ,  {\bf h}^{(2)} }  P({\bf v}, {\bf h}^{(1)} ,  {\bf h}^{(2)} ; \theta)  \nonumber \\
=& \frac{1}{\mathcal{Z}(\theta)} \sum_{{\bf h}^{(1)} ,  {\bf h}^{(2)} } \textrm{exp}  \big(  -E ({\bf v},{\bf h}; \theta)  \big)
\end{align}
where ${\mathcal{Z}(\theta)} $ is the partition function. The parameters in Eq. \ref{eq:dbm} can be initialized with a greedy layer-wise pretraining step \cite{Hinton06a}. Basically, it is to learn the current parameters of RBMs with CD, and the learned features of the current layer RBM are treated as the ``data" to train the next RBM in the stack. The joint representation learning of multiple inputs, as well as model parameter updating via mean-field will be introduced in the next part. 
\subsection{The description of joint representations}
Our K-fan deep model is a deep neural network with K different kinds of inputs coupled stochastic binary hidden units in a hierarchical structure. The inputs can be binary or real values, and they share a hidden layer via multi-layers non-linear transform of RBMs for each input. For clarity, we will use 3-way deep structure to explain our model, shown in Fig. \ref{fig:dbnkfan}. 

Suppose we have a set of visible inputs ${\bf v}^x \in \{0,1\}^D$, ${\bf v}^y \in \{0,1\}^D$ and ${\bf v}^z \in \{0,1\}^K$, and a sequence of layers of hidden units for each input. Note that ${\bf v}^x$, ${\bf v}^y$ and ${\bf v}^z$ can have different dimensionalities. For clarity, we start by modeling each input using a separate two-layer DBM. For input ${\bf v}^x$, and its two layers of  hidden units ${\bf h}^{(x1)} \in \{0,1\}^{n_{x1}}$ and ${\bf h}^{(x2)} \in \{0,1\}^{n_{x2}}$, the probability for the visible vector ${\bf v}^x$ is given by
\begin{align}\label{eq:dbmx}
&P({\bf v}^x; \theta_{x}) = \sum_{{\bf h}^{(x1)} ,  {\bf h}^{(x2)} }  P({\bf v}^x, {\bf h}^{(x1)} ,  {\bf h}^{(x2)} ; \theta_x)  \\
=& \frac{1}{\mathcal{Z}(\theta_x)} \sum_{{\bf h}^{(x1)} ,  {\bf h}^{(x2)} } \textrm{exp}  \bigg(   \sum_{k,j} W^{(x1)}_{k,j} v^x_{k} h^{(x1)}_j   + \sum_{j,l} W^{(x2)}_{j,l} h^{(x1)}_j h^{(x2)}_{l} \bigg)
\end{align}
where $\theta_x =\{ {\bf W}^{(x1)}, {\bf W}^{(x2)} \}$. Note that we only consider the binary observation (can be easily extended into real value case with Gaussian RBMs) and ignore biases for both visible and hidden units for clarity. Analogously, we can get the likelihoods for ${\bf v}^y$ and ${\bf v}^z$ respectively in the same formulas, but with different subscripts.  

To form our 3-fan deep model, we combine the three models from ${\bf v}^x$, ${\bf v}^y$ and ${\bf v}^z$, by adding an additional layer of binary
hidden units on top of them. The resulting graphical model is shown in the right panel of Fig. \ref{fig:dbnkfan}. The joint
distribution over the multi-modal inputs can be written as:
\begin{align}
& P({\bf v}^x,  {\bf v}^y, {\bf v}^z;  \theta) = \sum_{{\bf h}^{(x2)} ,  {\bf h}^{(y2)},  {\bf h}^{(z2)},  {\bf h}^{(3)}} P( {\bf h}^{(x2)} ,  {\bf h}^{(y2)},  {\bf h}^{(z2)}, {\bf h}^{(3)})   \nonumber \\
& \big(  \sum_{{\bf h}^{(x1)} } P({\bf v}^x,  {\bf h}^{(x1)},  {\bf h}^{(x2)}) \big)       \big( \sum_{{\bf h}^{(y1)} } P({\bf v}^y ,  {\bf h}^{(y1)},  {\bf h}^{(y2)})   \big)    \nonumber \\
& \big( \sum_{{\bf h}^{(z1)} } P({\bf v}^z ,  {\bf h}^{(z1)},  {\bf h}^{(z2)})   \big) 
\label{eq:jointprob}
\end{align}
where ${\bf W}^{(x3)}$,  ${\bf W}^{(y3)}$ and ${\bf W}^{(z3)}$ are respectively the top layer weights which connected to the top shared layer ${\bf h}^{(3)}$ for each input. And the parameters $\theta = \{ \theta_x, \theta_y, \theta_z,  {\bf W}^{(x3)},  {\bf W}^{(y3)}, {\bf W}^{(z3)}  \}$, where we ignore the biases. 

\subsection{ Learning and Inference}
In learning stage, all inputs are available. Thus, we can use CD to learn the shared presentation and model parameter effectively. In practice, we divide the parameter estimation into two stages: parameter initialization and fine-tuning. And the parameter initialization stage focuses on learning the joint representations shared by different modalities, while the fine-tuning stage emphasizes on the discriminative learning according to the properties of tasks in our hand. More specifically, because of the joint multimodal deep structure, we first initialize the model parameters by maximizing the joint likelihood in Eq. \ref{eq:jointprob}. Then we update our model parameters by optimizing different objective functions according to different tasks. Note that for different functions, we have the same parameter initialization step via CD because we use the same multimodal deep structure. 

\subsubsection{Parameter initialization}
We first use pretraining to initialize the weights of each branch separately. Basically, it is to learn a stack of RBMs greedily layer-by-layer. To put it simply, the learned features of the current layer RBM are treated as the ``data" to train the next RBM in the stack. Assume that we have a training set $\mathcal{D} = \langle {\bf v}_i^{x}, {\bf v}_i^{y}, {\bf v}_i^{z} \rangle_{i=1}^N$. Then, for each branch ${\bf v}^{x}$,  we can use RBMs to initialize the weights $\{ {\bf W}^{(x1)}, {\bf W}^{(x2)},  {\bf W}^{(x3)}\}$ in the layer-wise manner mentioned above. 

Then, we update the model parameters with CD. Because this model is intractable, we use an efficient approximate learning and inference  \cite{Salakhutdinov12}, such as mean-field method, to estimate data-dependent expectations, and an MCMC based stochastic approximation procedure to approximate the model's expected sufficient statistics. 
In variational learning \cite{Hinton93,Neal99,Salakhutdinov12} the true posterior distribution over latent variables $p({\bf h}|{\bf v}; \theta)$ for each training vector ${\bf v}$, is replaced by an approximate posterior $q({\bf h}|{\bf v}; \mu)$ and the parameters are updated by following the gradient of a lower bound on the log-likelihood:
\begin{align}
\textrm{ln} P({\bf v}; \theta) \ge \sum_{\bf h} q({\bf h} |{\bf v};\mu) \textrm{ln} p({\bf v},{\bf h}; \theta)  + \mathcal{H}(q) \\
=\textrm{ln} p({\bf v}; \theta) - KL[q({\bf h}|{\bf v};\mu)  || q({\bf h}|{\bf v};\mu)  ]
\end{align}
where ${\bf h} = \{  {\bf h}^{(x1)}, {\bf h}^{(x2)}, {\bf h}^{(y1)},  {\bf h}^{(y2)},  {\bf h}^{(z1)},  {\bf h}^{(z2)}, {\bf h}^{(3)}   \}$, ${\bf v} = \{ {\bf v}^x,  {\bf v}^y, {\bf v}^z \}$, and $\mu$ is the mean-field approximation to hidden variable (see further), and $\mathcal{H}(\cdot)$ is the entropy functional. To maximize the log-likelihood of the training data, is to find parameters that minimize the Kullback--Leibler divergences between the approximating and true posteriors. 

Similar to \cite{Srivastava14b}, we use the naive mean-field approach, with fully factorized distribution to approximate the true posterior:
\begin{align}
& q({\bf h}| {\bf v}; \mu) =\big( \prod_i q(h^{(3)}_i | {\bf v} \big)    \bigg( \prod_k q(h^{(x1)}_k | {\bf v}) \prod_j q(h^{(x2)}_j | {\bf v})  \bigg)    \nonumber \\
& \bigg( \prod_k q(h^{(y1)}_j | {\bf v}) \prod_j q(h^{(y2)}_j | {\bf v})  \bigg)   \bigg( \prod_k q(h^{(z1)}_k | {\bf v}) \prod_j q(h^{(z2)}_j | {\bf v})  \bigg)   
\end{align}
where $\mu  = \{ \mu^{(1)}_x,  \mu^{(2)}_x,   \mu^{(1)}_y,  \mu^{(2)}_y, \mu^{(1)}_z,  \mu^{(2)}_z,   \mu^{(3)}  \}$ are the mean-field parameters with 
\begin{align}
& q(h^{(xl)} = 1) = u^{(l)}_x, \textrm{for } l=1,2. \\
& q(h^{(yl)} = 1) = u^{(l)}_y, \textrm{for } l=1,2. \\
& q(h^{(zl)} = 1) = u^{(l)}_z, \textrm{for } l=1,2. \\
& q(h^{(3)} = 1) = u^{(3)}.
\end{align}
Then $\mu$ can be used to update the hidden variables in the data dependent item in Eq. \ref{eq:grad2}. And the model dependent hidden variables in Eq. \ref{eq:grad2} can be sampled with MCMC. Then, the model parameter can be updated with CD algorithm according to Eq. \ref{eq:grbm}.

\subsubsection{Parameter fine-tuning}
The parameter fine-tuning stage is different from previous methods, and is determined by the tasks that we pursue in our hand. In our experiments, we test our model on two kind of tasks: visual restoration and labeling, and multi-view multi-class object recognition. For the two tasks, we used the same deep structures with K=3, but with different objective functions. 

{\bf Joint visual restoration and labeling:} For the given corrupted input ${\bf v}^{x}$, we need to restore its clear image ${\bf v}^{y}$ as well as predict its label ${\bf v}^{z}$. Thus, we propose the following objective function for the binary case
\begin{align} 
& \theta  =\textrm{argmin}_{\theta} \mathcal{J}({\bf v}^{x}, {\bf v}^{y}, {\bf v}^{z}; \theta)  \nonumber \\
&= \textrm{argmin}_{\theta}   -  \sum_{i = 1}^N  {\bf v}_{i}^y \textrm{log} {\bf \hat{v}}_{i}^y +  (1- {\bf v}_{i}^y) \textrm{log} (1-{\bf \hat{v}}_{i}^y)  \nonumber \\
&  - \lambda \sum_{i = 1}^N  {\bf v}_{i}^z \textrm{log} {\bf \hat{v}}_{i}^z +  (1- {\bf v}_{i}^z) \textrm{log} (1-{\bf \hat{v}}_{i}^z) 
\label{eq:deepjoint}
\end{align}
where $\theta$ is the set of weights in the 3-way deep architecture respectively (we ignore the subscripts for clarity), and $\lambda$ is the constant to balance the two losses in Eq. \ref{eq:deepjoint}. And ${\bf \hat{v}}_i^y$ and ${\bf \hat{v}}_i^{z}$ are the predictions from the noise input ${\bf v}_{i}^x$, specified as follows
\begin{align} 
& {{\bf h}_i} = \underbrace{f_{L} \circ f_{L-1} \circ \cdot\cdot\cdot  \circ f_1}_{L \textrm{ times}}({\bf v}_{i}^x) \label{eq:eqhidden} \\
 & {\bf \hat{v}}_i^y  =   \underbrace{g_1 \circ g_2\circ \cdot\cdot\cdot \circ g_{L}}_{L\textrm{ times}}  ({{\bf h}_i}) \label{eq:eqpred} \\
 & {\bf \hat{v}}_i^z  =   \underbrace{\phi_1 \circ \phi_2\circ \cdot\cdot\cdot \circ \phi_{L}}_{L\textrm{ times}}  ({{\bf h}_i}) \label{eq:eqlabels}
\end{align}
where $\circ$ indicates function composition, ${\bf h}_i$ is the shared hidden representation from the triplet $\langle {\bf v}^{x}, {\bf v}^{y}, {\bf v}^{z}  \rangle$, and the functions $f_l$, $g_l$, and $\phi_l$ are non-linear projections, such as sigmoid function. We ignore the underscript for parameters in mapping functions $f_l$, $g_l$, and $\phi_l$, for $l = \{1, ..., L\}$ in the above equations. In our case, we use the same number of layers $L$ to all branches for simplicity and clarity. Note that each branch in the deep structure network can have different number of layers and nodes, only if they keep the same dimetionality of the shared representation. 

{\bf Multi-view multi-class object recognition:} Assume that ${\bf v}^{x}$ is the input image contains different objects, and ${\bf v}^{y}$ is the vector to describe the views to catch the object with cameras, and ${\bf v}^{z}$ indicates which class the object belongs to. The purpose is to answer whether these additional views with the low level image features are helpful to improve the recognition accuracy. Thus, we propose the following objective 
\begin{align}\label{eq:deepmult}
& \theta  =\textrm{argmin}_{\theta} \mathcal{L}({\bf v}^{x}, {\bf v}^{y}, {\bf v}^{z}; \theta)  \nonumber \\
&= \textrm{argmin}_{\theta} - \sum_{i = 1}^N  {\bf v}_{i}^z \textrm{log} {\bf \hat{v}}_{i}^z +  (1- {\bf v}_{i}^z) \textrm{log} (1-{\bf \hat{v}}_{i}^z) 
\end{align}
where $\theta$ is the set of the weights in the 3-way deep architecture as before. And ${\bf \hat{v}}_i^{z}$ is the prediction from the image ${\bf v}_{i}^x$ and its view ${\bf {v}}_i^y$, specified as follows
\begin{align} 
& {{\bf h}^{(x2)}} = \underbrace{f_{L-1} \circ \cdot\cdot\cdot  \circ f_1}_{L-1 \textrm{ times}}({\bf v}_{i}^x) \label{eq:eqhidden1}  \\
& {{\bf h}^{(y2)}} = \underbrace{g_{L-1} \circ \cdot\cdot\cdot  \circ g_1}_{L-1 \textrm{ times}}({\bf v}_{i}^y) \label{eq:eqhidden2} \\
 & {\bf h}_i  =   \textrm{sigmoid}  ( {{\bf h}^{(x2)}}^T {\bf W}^{(x3)}  +   {{\bf h}^{(y2)}}^T {\bf W}^{(y3)}    ) \label{eq:eqhidden} \\
 & {\bf \hat{v}}_i^z  =   \underbrace{\phi_1 \circ \phi_2\circ \cdot\cdot\cdot \circ \phi_{L}}_{L\textrm{ times}}  ({{\bf h}_i}) \label{eq:eqlabels}
\end{align}
where we ignore the bias term for ${\bf h}_i$ for clarity. 

Given the parameter initialization with CD algorithm, we can minimize the objective function in Eq. \ref{eq:deepjoint} or \ref{eq:deepmult} respectively to estimate the model parameters. To fine-tune the model, we compute the gradients w.r.t. weights via backpropagation in each layer in the objective function, and then we use any gradient-based methods to update the model parameters, such as L-BFGS \cite{Byrd95}. 

Note that the joint visual restoration and labeling task is different from the multi-view multi-class recognition task. In fact, we can see the differences between the two objective functions in Eqs. \ref{eq:deepmult} and \ref{eq:deepjoint}, even though they used the same multimodal deep structure and initialized the model parameters with the same CD algorithm. The multi-view multi-class object recognition leverage multiple inputs to improve the classification performance, while the joint visual restoration and labeling learns a joint model for multiple outputs from only one input channel. 
The former has one input and two outputs, thus it is related to multi-task learning. While the latter has two inputs and one output prediction, by leveraging multi-source information for multi-classification problem.

\subsection{Relationship to other models}
We analyzed the differences between our model and other deep structures, such as deep autoencoder and deep Boltzmann machines. 
\subsubsection{deep autoencoder}
The deep autoendcoder \cite{Hinton06b} can be thought as a bi-modal deep model with feed-forward networks. It consists of encoder and decoder in order to recover the data itself by learning the shared hidden representation. On the contrary, our model is a K-way deep feed-forward neural network with multiple channels and our model can adjust to different optimization problems given the same architecture, for example the joint visual restoration and labeling. Although these two models can use the same CD algorithm to initialize model parameters, our model is more powerful to handle multiple output predictions, instead of just recovering the data in the deep autoencoder. 

\subsubsection{deep Boltzmann machines}
A multimodal DBM can be viewed as a composition of unimodal undirected pathways. Each path-way can be pretrained separately in a completely unsupervised fashion, which make it possible to leverage a large supply of unlabeled data. 
The middle graph in Fig. \ref{fig:dbnkfan} is a 3-fan multimodal DBM. In our framework, we define a K-fan deep structure, where each branch can be a deep learning model, such as DBN, LSTM and CNN for different multimodal inputs. Thus, our model is more powerful. Moreover, the multimodal DBM is a generative model, while our model is a discriminative model. In other words, our modal is a multi-path feed-forward neural network with a shared representation, which can be optimized
in an discriminative manner to handle different tasks. 

\subsubsection{Multi-task learning}
Multi-task learning is an approach to learn a problem together with other related problems at the same time, using a shared representation. Our multiple deep neural network can predict multiple outputs by learning a shared representation for multi-task. Thus, our model can be used to handle multi-task learning problems, such as the joint visual restoration and labeling. In addition, our model can leverage multiple inputs or resources to improve the prediction or classification, such as the multi-view multi-class object recognition in our case. Thus, our deep K-fan structure is flexible and powerful, and can be optimized according to different tasks.

\section{Experiments}
As mentioned before, we test our deep model with two tasks: joint visual restoration and labeling, and multi-view multi-class object recognition. For the former task, we evaluated the performance with Peak signal-to-noise ratio (PSNR). In addition, we also used error rate to evaluate whether denoising is helpful or not in the recognition task. As for the multi-view multi-class object recognition, we leverage the lower level image features with additional multiple sources, such as camera views to improve the classification accuracy. 

\subsection{Data description}
{\bf The MNIST dataset}\footnote{\url{http://yann.lecun.com/exdb/mnist/}} consists of $28\times28$-size images of handwriting digits from $0$ through $9$ with a training set of 60,000 examples and a testing set of 10,000 examples, and has been widely used to test character denoising and recognition methods. A set of examples are shown in Fig. \ref{fig:denoised}(a). In the experiment, we test both denoising and recognition performance. As for the noise model, we considered the structural noise that is hard to remove in the handwriting images. Basically,  we random add two strokes to each digits, refer the structural noise in Fig. \ref{fig:denoised}(b), which shows heavily corrupt images, with more than 50\% regions. 

{\bf The USPS Handwritten binary Alphadigits}\footnote{\url{http://www.cs.nyu.edu/~roweis/data/binaryalphadigs.mat}} are binary images with size $20 \times16$ pixels. There are digits of ``0" through ``9" and capital ``A" through ``Z", with 39 examples of each class. In our experiments, we only test our method on the binary Alphabets.

{\bf The multi-view multi-class dataset}\footnote{\url{http://cvlab.epfl.ch/data/multiclass}} consists of 23 minutes and 57 seconds of synchronized frames taken at 25fps from 6 different calibrated DV cameras. One camera was placed about 2m high of the ground, two others where located on a first floor high, and the rest on a second floor to cover an area of $22m \times 22m$. This ground truth contains 242 annotated multi-view non-consecutive frames. These frames contain different real situations where pedestrians, cars and buses appear and can cause high occlusions among them. 
In our task, we want to test whether the additional multi-view information is helpful or not in the multi-classification task. Hence, we crop all objects in the frames according to its groundtruth bounding box, and then we resize them into the same size $88\times64$. Then, we vectorize all the cropped objects into 5632 dimensions, with additional 6 dimension camera view information to each object. Finally, we get 4907 instances, with 1295 persons, 3354 cars and 58 buses.

\subsection{Experimental setting}
In all experiments, we use the 3-fan deep model with 3-layer nonlinear mapping in each pathway, with the learning rate 0.1 and CD-1 step sampling to initialize the model weights. We set the constant $\lambda=1$ to weigh the two term in the objective function in Eq. \ref{eq:deepjoint}. In the fine-tuning state, we used L-BFGS to update the model parameters. In particular, for the joint restoration and labeling, we used CD (multimodal DBMs) to initialize the weights, and then optimize Eq. \ref{eq:deepjoint} to fine-tune the model parameters with L-BFGS. For the MNIST digits and the USPS alphabets, we set the number of hidden nodes [400 200 250] for each pathway respectively in the 3-layer deep model.  

As for the multi-view multi-class object recognition task, we used the same CD to initialize the weights by maximizing the joint likelihood, and then optimize Eq. \ref{eq:deepmult} to fine-tune the model parameters. And we evaluate the performance with 10-fold cross validation. As for the model structure, we set the number of hidden nodes [400 200 250] in each pathway respectively in the 3-layer deep model. 

\subsection{Experimental results}
We evaluated our method on both joint visual restoration and labeling and multi-view multi-class object recognition tasks.
\subsubsection{Joint restoration and labeling}
In this part, we tested our model on the multi-task learning problem: joint restoration and labeling by minimizing Eq. \ref{eq:deepjoint}. Before analyzing the performance, we first evaluated the PSNR lower bound on the noise images, as well as the accuracy upper bound on the clear images. 

{\bf PSNR and error rate bounds}: 
(1) generating the noisy digits: we added noise to each MNIST image by randomly drawing two strokes to construct its noisy observation (clutter more than 50\% regions). Thus, for all the clear training and testing digits, with 60,000 training and 10,000 testing images respectively, we can construct the corresponding noisy images. 
(2) training a classifier on the clear data: we first learned the deep neural network (DNN) \cite{Hinton06a} for classification tasks by minimizing the cross entropy. In the experiment on the MNIST digits, we used the default DDN parameters, namely 4-layer deep structure with hidden nodes [500 500 2000 10] respectively for each layer. Then we trained the model on the 60,000 clean images to learn the DNN classifier and then tested it on the 10,000 clean testing set. The error rate on the clean testing set we can get using DNN is 1.2\%, while 
the error rates on the noisy testing set is 61.0\% heavily corrupted noise in Fig. \ref{fig:denoised}(b). And the PSNR lower bound on the noisy digits is 7.65 dB, which is calculated on the noisy testing set. 

{\bf Evaluation}:
To test our joint learning model, we trained our model on the 60,000 triplets: clean digit, noisy digit and its label. Then, we tested it on the 10,000 noisy testing dataset to restore its clear image and also predict its label. The denoised result (random sampled) of our model was shown in Fig. \ref{fig:denoised}(d), and its quantitative result was shown in Table \ref{tab:mnist}.

We compared our method to competitive baselines, which has separated pipelines, restoring image first and then recognizing it. Note that all the quantitative results of baselines are evaluated on the denoised images using the DNN classifier. More specifically, for each baseline, we first use it to denoise images, and then we use the DNN classifier to evaluate the recognition accuracy on the denoised images. The denoising result with denoising autoencoder (DAE)\cite{Vincent10} is shown in Fig. \ref{fig:denoised}(c), and its recognition accuracy is shown in Table \ref{tab:mnist}. And it clearly demonstrates that our joint learning method is better than DAE and ROBM \cite{Tang12}. 

We also evaluated our method on the USPS alphabets. Similar to the experiment on the MNIST digits, we added random strokes to the alphabets to create the noisy observations. Because there are only 39 training images for each class, we generated 10 corrupted samples for each clean image. In the experiments, we divided the total $39\times26$ binary images into training set (accounting for 80\%) and testing set (the rest 20\%), and we learned a 2-layer DNN model with 100 and 64 hidden nodes in each layer respectively on the clean training set. The classification performance on the clean testing set is 1.29\% in Table \ref{tab:usps}, while the error rate on the corresponding noisy testing set is 67.4\%. Then we used the learned DNN model to evaluate the denoising performance for the baselines. The visual performance of denoise autoencoder is shown in Fig. \ref{fig:usps}(c), and the result of our model is shown in Fig. \ref{fig:usps}(d). The quantitative comparison between our method and the baselines is shown in Table \ref{tab:usps}, which demonstrates our method for joint visual restoration and labeling is better than competitive baselines on both denoising and recognition tasks.


\begin{figure*}[t!]
\centering
\begin{tabular}{cc}
\includegraphics[trim = 48mm 99mm 45mm 95mm, clip, width=8.4cm]{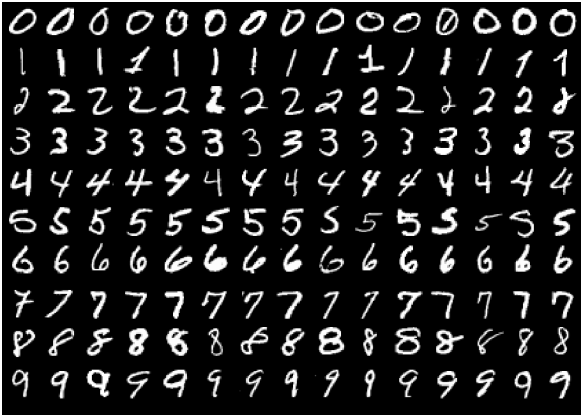}&
\includegraphics[trim = 48mm 99mm 45mm 95mm, clip, width=8.4cm]{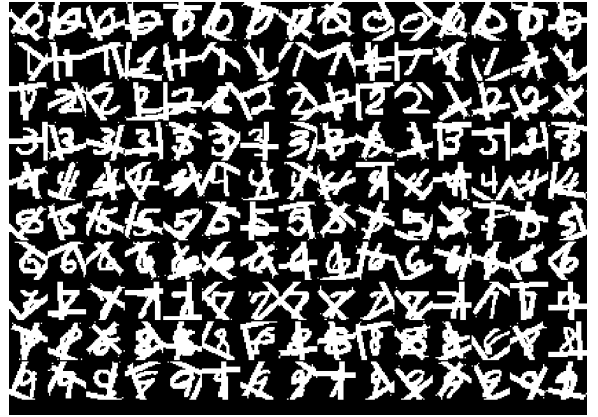} \\
(a) & (b) \\
\includegraphics[trim = 48mm 99mm 45mm 95mm, clip, width=8.4cm]{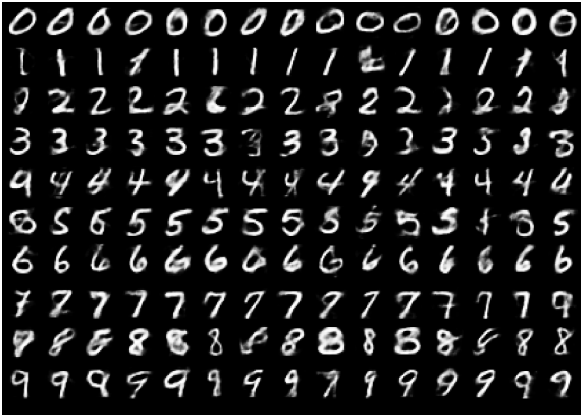} &
\includegraphics[trim = 48mm 99mm 45mm 95mm, clip, width=8.4cm]{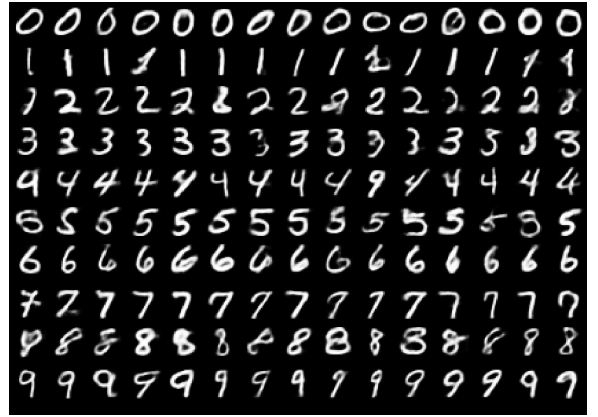} \\ 
(c) & (d)
\end{tabular}
\caption{The denoising results comparison on the heavily occluded MNIST digits. (a) original images; (b) noisy images with random structures; (c) denoising results with denoising autoencoder; (d) denoising results with our joint restoration and labeling model.} 
\label{fig:denoised}
\end{figure*}

\begin{table}[t!]
\centering
\begin{tabular}{lrr}
\hline
Model & PSNR (dB) & Error rate (\%) \\
\hline
Wiener \cite{Wiener64} & 11.7 & 58.5\\
RoBM \cite{Tang12} & 13.9 & 52.6\\
DAE & 13.58 &  35.9 \\
Our method & {\bf 18.6}&  {\bf 12.7}\\
\hline
DNN \cite{Hinton06b}& $\geq7.65$ & $1.20 \sim 61.0$\\
\hline
\end{tabular}
\caption{The experimental comparison on the heavily corrupted MNIST digits. It indicates that our method is better than competitive baselines on both denoising and recognition tasks.}
\label{tab:mnist}
\end{table}

%
\begin{figure*}[t!]
\centering
\begin{tabular}{cccc}
\includegraphics[trim = 76mm 73.5mm 106.5mm 60mm, clip, width=3.6cm]{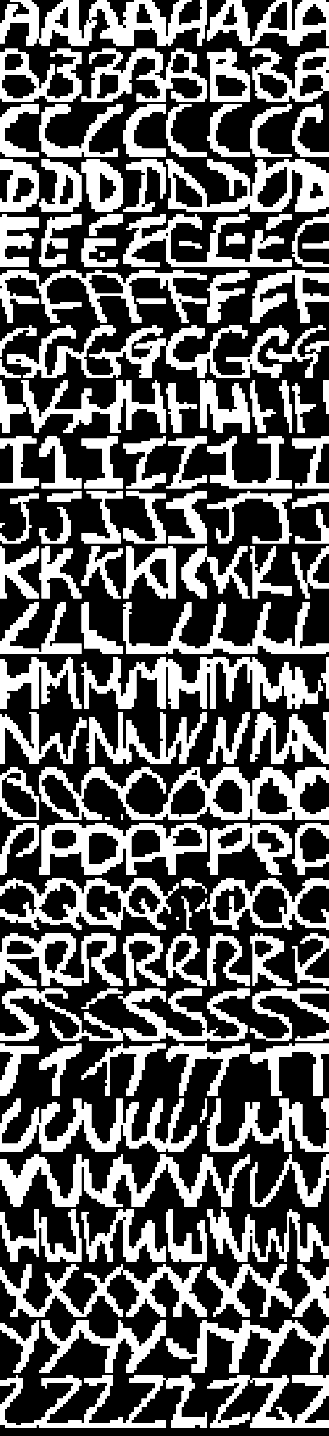}&
\includegraphics[trim = 76mm 73.5mm 106.5mm 60mm, clip, width=3.6cm]{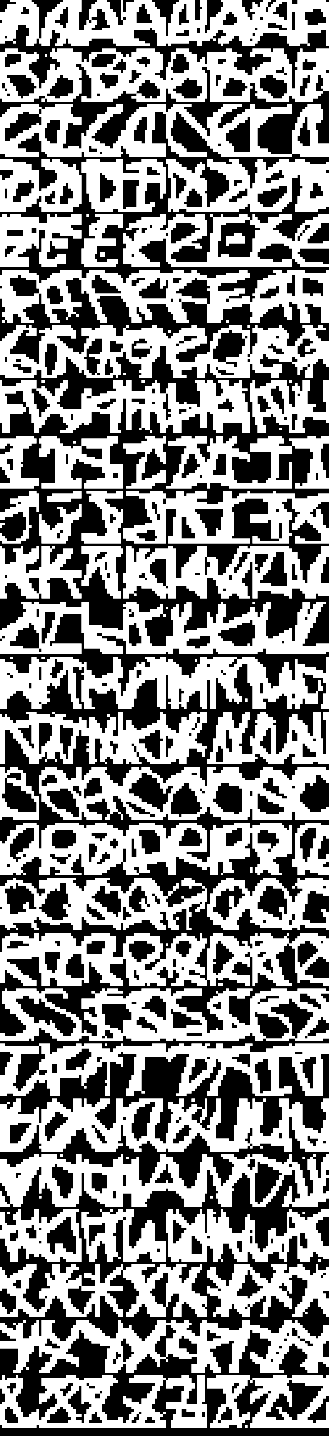} &
\includegraphics[trim = 76mm 73.5mm 106.5mm 60mm, clip, width=3.6cm]{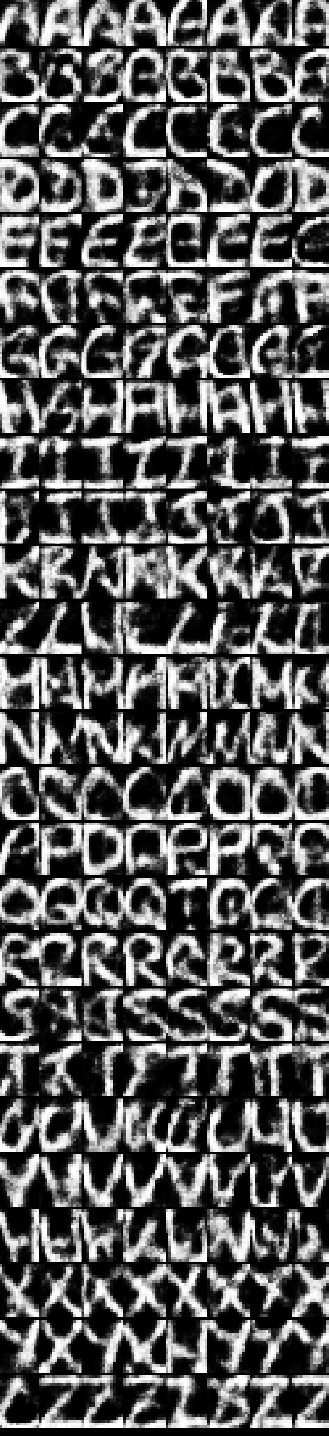} &
\includegraphics[trim = 76mm 73.5mm 106.5mm 60mm, clip, width=3.6cm]{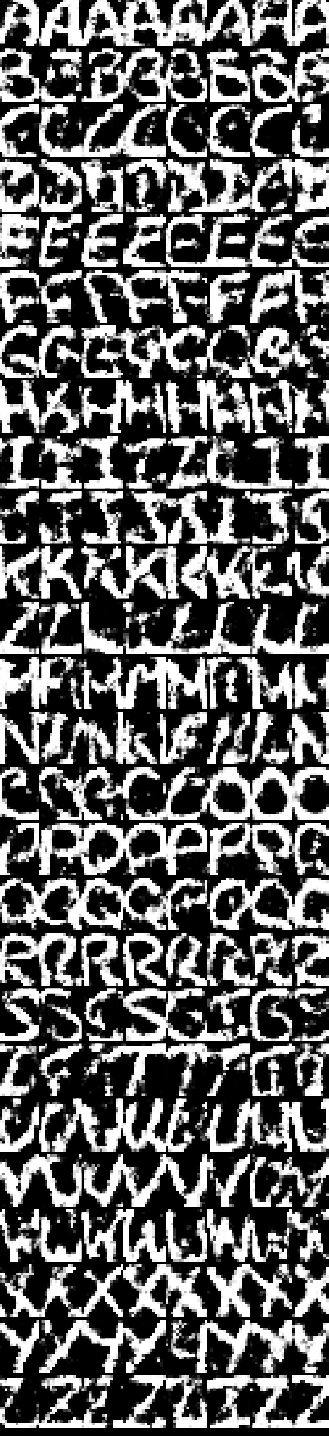} \\
(a) & (b) & (c) &(d)
\end{tabular}
\caption{The denoising results comparison on USPS alphabet (the type 2 noise). (a) original images from `A' to `Z' arranged in the top-down manner; (b) noisy images with random structures; (c) denoising results with denoising autoencoder; (d) denoising results with our 3-fan deep model. }
\label{fig:usps}
\end{figure*}

\begin{table}[h!]
\centering
\begin{tabular}{lrr}
\hline
Model & PSNR (dB) & Error rate (\%) \\
\hline
Wiener \cite{Wiener64} & 14.2 & 67.8 \\
RoBM \cite{Tang12} & 16.3 & 62.8\\
DAE & 19.2&  42.5 \\
Our method & {\bf 19.6} &  {\bf 32.8}\\
\hline
DNN \cite{Hinton06b}& $\geq 8.12$ & $1.29 \sim 67.4$\\
\hline 
\end{tabular}
\caption{The experimental comparison on USPS alphabets. The PSNR value of DNN is 8.12 dB, which shows the lower bound on the noisy testing set. The error rate using DNN means that the error rates on the noisy testing set and the original clean testing set are 67.4\% and 1.29\% respectively. It demonstrates that our method is better than competitive baselines on both denoising and recognition tasks.}
\label{tab:usps}
\end{table}

%

\subsubsection{Multi-view multi-class recognition}
In this part, we test our model on the multi-view multi-class recognition task. In this task, our model has two inputs and one output prediction, thus it can leverage multiple sources to boost classification accuracy. 

The multi-view multi-class dataset contains total $4907\times 5632$ instances, belonging to 3 classes. And there are additional 6 bits camera view information for each instance. This multi-class dataset is unbalanced with 1295 persons, 3354 cars and 58 buses. The purpose is to answer whether the additional camera information is helpful or not in the multi-classification tasks. In our experiment, we use 10-fold cross validation, by randomly sampling 9 fold for training and the rest for testing. To train our 3-fan multimodal deep model, we can leverage both lower level image features and multi-view information as two inputs, and predict the object class in the output. We optimize our model via Eq. \ref{eq:deepmult} and show our result in Table \ref{tab:multi-sources}. The SVM with multi-view yields better result than SVM w/o multi-view, which indicates that the additional multi-view information is helpful. And our model outperforms multimodal DBM and SVM, which clearly demonstrates that our 3-fan deep model can effective leverage multiple sources to boost performance. It also indicates our discriminative model is better than multimodal DBM in the classification problem.

To sum up, we can optimize different objective functions to achieve different goals. For example, if we want to handle multi-task learning, we can use the objective function in Eq. \ref{eq:deepjoint} for multiple outputs. If we want to leverage multiple sources to boost classification performance, we can propose a similar objective function as in Eq. \ref{eq:deepmult}. However, no matter what objective function we use, we still use the same K-fan deep multimodal structure. Thus, our model is flexible and powerful to leverage multiple inputs for multiple predictions.

\begin{table}[t!]
\centering
\begin{tabular}{lr}
\hline
Model &  Error rate (\%) \\
\hline
SVM w/o multi-view & 9.59\\
SVM w multi-view & 6.92\\
Multimodal DBM \cite{Srivastava14b} & 7.10 \\
Our model & {\bf 4.80}\\
\hline
\end{tabular}
\caption{The experimental comparison on the multi-view multi-class recognition dataset. It demonstrates that our model outperforms other methods significantly.}
\label{tab:multi-sources}
\end{table}


\section{Conclusions}
In this paper, we propose a generalized K-fan deep structure, with shared representations for multimodal learning. Our deep multimodal structure is powerful because each branch can be a deep learning model for different inputs. Given the deep model, we can optimize different objective functions to learn the shared representation from multimodalities to handle different tasks. To learn the model parameters, we take a two stage steps: parameter initialization and parameter fine-tuning. In the parameter initialization stage, we use the CD algorithm to maximize the joint likelihood as the multimodal DBM does. In the fine-tuning stage, we update the model parameter according to the defined objective function. We test our model on two tasks: the joint restoration and labeling, and the multi-view multi-class object recognition. The former task is to handle the multi-task learning problem, while the latter is to answer whether our model can leverage multiple sources to boost classification performance. 
The experimental results show our K-fan deep structure model is flexible and powerful, and can be optimized according to different functions to address different problems.




%
\bibliographystyle{IEEEtran}
\bibliography{crbmbib}  

\begin{thebibliography}{10}
\providecommand{\url}[1]{#1}
\csname url@samestyle\endcsname
\providecommand{\newblock}{\relax}
\providecommand{\bibinfo}[2]{#2}
\providecommand{\BIBentrySTDinterwordspacing}{\spaceskip=0pt\relax}
\providecommand{\BIBentryALTinterwordstretchfactor}{4}
\providecommand{\BIBentryALTinterwordspacing}{\spaceskip=\fontdimen2\font plus
\BIBentryALTinterwordstretchfactor\fontdimen3\font minus
  \fontdimen4\font\relax}
\providecommand{\BIBforeignlanguage}[2]{{%
\expandafter\ifx\csname l@#1\endcsname\relax
\typeout{** WARNING: IEEEtran.bst: No hyphenation pattern has been}%
\typeout{** loaded for the language `#1'. Using the pattern for}%
\typeout{** the default language instead.}%
\else
\language=\csname l@#1\endcsname
\fi
#2}}
\providecommand{\BIBdecl}{\relax}
\BIBdecl

\bibitem{Hinton06b}
G.~E. Hinton and R.~R. Salakhutdinov, ``Reducing the dimensionality of data
  with neural networks,'' \emph{Science}, vol. 313, no. 5786, pp. 504--507,
  Jul. 2006.

\bibitem{Ngiam10}
J.~Ngiam, A.~Khosla, M.~Kim, J.~Nam, H.~Lee, and A.~Y. Ng, ``Multimodal deep
  learning.'' in \emph{ICML}, 2011, pp. 689--696.

\bibitem{Hinton06a}
\BIBentryALTinterwordspacing
G.~E. Hinton, S.~Osindero, and Y.-W. Teh, ``A fast learning algorithm for deep
  belief nets,'' \emph{Neural Comput.}, vol.~18, no.~7, pp. 1527--1554, Jul.
  2006. [Online]. Available:
  \url{http://dx.doi.org/10.1162/neco.2006.18.7.1527}
\BIBentrySTDinterwordspacing

\bibitem{Bengio12}
Y.~Bengio, A.~Courville, and P.~Vincent, ``Representation learning: A review
  and new perspectives,'' \emph{TPAMI}, 2012.

\bibitem{Srivastava14b}
N.~Srivastava and R.~Salakhutdinov, ``Multimodal learning with deep boltzmann
  machines,'' \emph{Journal of Machine Learning Research}, vol.~15, pp.
  2949--2980, 2014.

\bibitem{Fukushima80}
K.~Fukushima, ``{N}eocognitron: {A} self-organizing neural network model for a
  mechanism of pattern recognition unaffected by shift in position,''
  \emph{Biological Cybernetics}, vol.~36, pp. 193--202, 1980.

\bibitem{LeCun89}
Y.~LeCun, B.~Boser, J.~S. Denker, D.~Henderson, R.~E. Howard, W.~Hubbard, and
  L.~D. Jackel, ``Backpropagation applied to handwritten zip code
  recognition,'' \emph{Neural Comput.}, vol.~1, no.~4, pp. 541--551, Dec. 1989.

\bibitem{Hochreiter97}
S.~Hochreiter and J.~Schmidhuber, ``Long short-term memory,'' \emph{Neural
  Comput.}, vol.~9, no.~8, pp. 1735--1780, Nov. 1997.

\bibitem{Xing05}
E.~P. Xing, R.~Yan, and A.~G. Hauptmann, ``Mining associated text and images
  with dual-wing harmoniums,'' in \emph{In Conference on Uncertainty in
  Artificial Intelligence}, 2005.

\bibitem{Huiskes10}
\BIBentryALTinterwordspacing
M.~J. Huiskes, B.~Thomee, and M.~S. Lew, ``New trends and ideas in visual
  concept detection: The mir flickr retrieval evaluation initiative,'' in
  \emph{Proceedings of the International Conference on Multimedia Information
  Retrieval}, ser. MIR '10.\hskip 1em plus 0.5em minus 0.4em\relax New York,
  NY, USA: ACM, 2010, pp. 527--536. [Online]. Available:
  \url{http://doi.acm.org/10.1145/1743384.1743475}
\BIBentrySTDinterwordspacing

\bibitem{Guillaumin10}
M.~Guillaumin, J.~J. Verbeek, and C.~Schmid, ``Multimodal semi-supervised
  learning for image classification.'' in \emph{CVPR}.\hskip 1em plus 0.5em
  minus 0.4em\relax IEEE, 2010, pp. 902--909.

\bibitem{Vincent10}
P.~Vincent, H.~Larochelle, I.~Lajoie, Y.~Bengio, and P.-A. Manzagol, ``Stacked
  denoising autoencoders: Learning useful representations in a deep network
  with a local denoising criterion,'' \emph{JMLR}, 2010.

\bibitem{Holmstrom90}
L.~Holmstr{\"o}m and P.~Koistinen, ``Using additive noise in back-propagation
  training,'' Rolf Nevanlinna Institute, Research Reports~A3, 1990.

\bibitem{Tang12}
Y.~Tang, R.~Salakhutdinov, and G.~Hinton, ``Robust boltzmann machines for
  recognition and denoising,'' in \emph{CVPR}.\hskip 1em plus 0.5em minus
  0.4em\relax Washington, DC, USA: IEEE Computer Society, 2012, pp. 2264--2271.

\bibitem{Torralba07}
\BIBentryALTinterwordspacing
A.~Torralba, K.~P. Murphy, and W.~T. Freeman, ``Sharing visual features for
  multiclass and multiview object detection,'' \emph{IEEE Trans. Pattern Anal.
  Mach. Intell.}, vol.~29, no.~5, pp. 854--869, May 2007. [Online]. Available:
  \url{http://dx.doi.org/10.1109/TPAMI.2007.1055}
\BIBentrySTDinterwordspacing

\bibitem{Salakhutdinov12}
R.~Salakhutdinov and G.~E. Hinton, ``An efficient learning procedure for deep
  boltzmann machines,'' \emph{Neural Computation}, vol.~24, no.~8, pp.
  1967--2006, 2012.

\bibitem{Hinton93}
G.~E. Hinton and R.~S. Zemel, ``Autoencoders, minimum description length and
  helmholtz free energy.'' in \emph{NIPS}, J.~D. Cowan, G.~Tesauro, and
  J.~Alspector, Eds.\hskip 1em plus 0.5em minus 0.4em\relax Morgan Kaufmann,
  1993, pp. 3--10.

\bibitem{Neal99}
R.~M. Neal and G.~E. Hinton, ``Learning in graphical models,'' M.~I. Jordan,
  Ed.\hskip 1em plus 0.5em minus 0.4em\relax Cambridge, MA, USA: MIT Press,
  1999, ch. A View of the EM Algorithm That Justifies Incremental, Sparse, and
  Other Variants, pp. 355--368.

\bibitem{Byrd95}
\BIBentryALTinterwordspacing
R.~H. Byrd, P.~Lu, J.~Nocedal, and C.~Zhu, ``A limited memory algorithm for
  bound constrained optimization,'' \emph{SIAM J. Sci. Comput.}, vol.~16,
  no.~5, pp. 1190--1208, Sep. 1995. [Online]. Available:
  \url{http://dx.doi.org/10.1137/0916069}
\BIBentrySTDinterwordspacing

\bibitem{Wiener64}
N.~Wiener, \emph{Extrapolation, Interpolation, and Smoothing of Stationary Time
  Series}.\hskip 1em plus 0.5em minus 0.4em\relax The MIT Press, 1964.

\end{thebibliography}
\end{document}